# Is Your Model "MADD"? A Novel Metric to Evaluate Algorithmic Fairness for Predictive Student Models

Mélina Verger
Sorbonne Université, CNRS,
LIP6, F-75005 Paris, France
melina.verger@lip6.fr

Sébastien Lallé
Sorbonne Université, CNRS,
LIP6, F-75005 Paris, France
sebastien.lalle@lip6.fr

François Bouchet
Sorbonne Université, CNRS,
LIP6, F-75005 Paris, France
francois.bouchet@lip6.fr

Vanda Luengo
Sorbonne Université, CNRS,
LIP6, F-75005 Paris, France
vanda.luengo@lip6.fr

## ABSTRACT
Predictive student models are increasingly used in learning environments due to their ability to enhance educational outcomes and support stakeholders in making informed decisions. However, predictive models can be biased and produce unfair outcomes, leading to potential discrimination against some students and possible harmful long-term implications. This has prompted research on fairness metrics meant to capture and quantify such biases. Nonetheless, so far, existing fairness metrics used in education are predictive performance-oriented, focusing on assessing biased outcomes across groups of students, without considering the behaviors of the models nor the severity of the biases in the outcomes. Therefore, we propose a novel metric, the Model Absolute Density Distance (MADD), to analyze models' discriminatory behaviors independently from their predictive performance. We also provide a complementary visualization-based analysis to enable fine-grained human assessment of how the models discriminate between groups of students. We evaluate our approach on the common task of predicting student success in online courses, using several common predictive classification models on an open educational dataset. We also compare our metric to the only predictive performance-oriented fairness metric developed in education, ABROCA. Results on this dataset show that: (1) fair predictive performance does not guarantee fair models' behaviors and thus fair outcomes, (2) there is no direct relationship between data bias and predictive performance bias nor discriminatory behaviors bias, and (3) trained on the same data, models exhibit different discriminatory behaviors, according to different sensitive features too. We thus recommend using the MADD on models that show satisfying predictive performance, to gain a finer-grained understanding on how they behave and regarding who and to refine models selection and their usage. Altogether, this work contributes to advancing the research on fair student models in education. Source code and data are in open access at https://github.com/melinaverger/MADD.

## 1. INTRODUCTION
Over the past decade, extensive research has focused on predictive student modeling for educational applications. The systematic literature review of Hellas et al. [15] has identified no less than 357 relevant papers on the matter published between 2010 and mid-2018. One of the most popular modeling technique in these works are machine learning (ML) classifiers, as many important predictive tasks in education can be framed as binary classification problems, e.g. to predict dropout, course completion, university admission, scholarship awarding. These classification models have thus gained widespread adoption, and the multiple stakeholders involved in education have recognized their potential to improve student learning outcomes and experience [29, 16].

However, in recent years, there have been concerns about the fairness of the models (also called *algorithmic fairness* [3]) used in education [3, 20, 11, 33]. This stems from a more general trend of research in ML and Artificial Intelligence (AI), where a large body of research has shown that classifiers, and AI models in general, can produce biased and unfair outcomes, e.g. [27, 5, 4, 23, 10]. This has led to increased public awareness about the potential harms of AI predictive models and the enforcement of stricter regulations[1]. In education too, recent studies have found that classification-based student models can be biased against certain groups of students, which could in turn significantly hinder their learning experience and academic achievements [3, 20, 33, 14, 17, 25].

---

---

[1] e.g. General Data Protection Regulation (2016) at European level, California Consumer Privacy Act (2018) at the United-States level, Principles on Artificial Intelligence (2019) from OECD (Organization for Economic Cooperation and Development) at the international level, and more specifically the upcoming European AI Act [32].

To unveil, measure and mitigate algorithmic unfairness, recent literature in AI has seen a proliferation of *fairness metrics* [35, 7]. Although many types of metrics exist (see Section 2), some of them require extensive prior knowledge and in practice the most common fairness metrics used in AI are statistical [35]. Statistical metrics aim at quantifying the differences in performance of a set of classification models across different groups of interest, with the assumption that fair classifiers should achieve similar performance across groups [7]. This is especially meaningful when some of the groups are known to be vulnerable to unfair model predictions. For instance, students with disability might be unfairly classified as at-risk of dropping out of an online course because the features used to train the classifiers did not capture well the different way they engage with the learning material [13] – when they can interact at all, since many K-12 material or educational technologies remain inaccessible [31, 6]. Hovewer, the pitfall of the existing statistical metrics is that they are all *predictive performance-oriented*, meaning that they solely consider the predictive performance of the classification model across predefined groups, disregarding that two classifiers with equal predictive performance can exhibit very different, and possibly unfair, behaviors. In particular, a classifier could produce similar error rates across two groups, but the actual errors made could be substantially more harmful to one of the group than the other.

In this paper, we thus propose a new statistical metric, the *Model Absolute Density Distance* (MADD), to analyze a model's discriminatory behaviors independently from its predictive performance. We also propose a complementary visualization-based analysis, which allows to inspect and qualify the models' discriminatory behaviors uncovered by the MADD. Altogether, this makes it possible to not only quantify, but also understand in a fine-grained way whether and how a given classifier may behave differently between the groups. As a case study, we apply our approach on the common task of predicting student success to a course, on open data for the sake of replication and on four common predictive classification models for the sake of generalization. We also compare our metric to ABROCA (*Absolute Between-ROC Area*), the only predictive performance-oriented metric developed in education [14] to the best of our knowledge. This case study shows that the MADD can successfully capture fine-grained models' discriminatory behaviors.

The remainder of this paper is organized as follows. Section 2 reports on related work on fairness metrics and their usage in education. Section 3 presents the MADD metric and the visualization-based analysis we propose to inspect and characterize models' discriminatory behaviors. Section 4 describes the experimental setup with which we applied our proposed approach in order to demonstrate its benefits. Section 5 presents our results and our comparison with ABROCA. In Section 6, we discuss more generally what our approach allows to unveil, the strengths and limitations it currently has as well as some practical guidelines, before concluding in Section 7 with future work.

## 2. RELATED WORK

Several fairness metrics have been proposed in AI for classification models. These metrics mostly fall into three categories: counterfactual (or causality-based), similarity-based (or individual), and statistical (or group) [35]. The first two categories, counterfactual and similarity-based, are seldom used in practice because they require extensive prior knowledge. More precisely, counterfactual metrics require building a directed acyclic causal graph with the nodes representing the features of an applicant and the edges representing relationships between the features [35]. Generating such a causal graph is typically not feasible without extensive studies to formally identify these relations. Similarity-based metrics require defining *a priori* a distance metric to measure how "similar" two individuals are, as well as to know from which value the models' results are considered "dissimilar" enough for these two individuals to be pointed out as unfairness. In contrast, statistical metrics, the category into which MADD falls, are easier to implement and more popular, as they solely require to identify *a priori* the groups of persons who might suffer from unfair classifications. As noted in the introduction, these metrics have so far sought to quantify differences in classification performance across the groups, and thus can be considered *predictive performance-oriented* only. However, a classifier that has similar error rates across two groups might actually produce errors that are harmless to a group but very harmful to the other, an aspect that is not quantified by existing statistical metrics. In this paper, we focus on the new MADD metric meant to assess unfair behaviors of classifiers, independently from their predictive performance. We recommend using it as a complement to a predictive performance analysis, rather than using predictive performance-oriented fairness metrics only, in order to gain a more refined and comprehensive understanding of the classifiers fairness.

In education, fairness studies are more recent and sparse (see overview in [3, 20]), and only a handful of them have focused on the fairness of classification models used in this context [14, 17, 18, 30, 25]. In Gardner et al. [14], the authors propose a new predictive performance-oriented fairness metric based on the comparison of the Areas Under the Curve (AUC) of a given predictive model for different groups of students. They used their metric to assess gender-based (male vs. female) differences in classification performance of MOOC dropout models, showing that ABROCA can capture unfair classification performance related to the gender imbalance in the data. ABROCA was also used in other educational studies, to evaluate the fairness across different sociodemographic groups of classifiers meant to predict college graduation [18], and categorize students' educational forum posts [30]. The other fairness studies in education have used more common statistical metrics in AI, such as group fairness, equalized odds, equal opportunity, true positive rate and false positive rate between groups, to predict course completion [26], at-risk students [17], and college grades and success [19, 36, 25]. Similarly to ABROCA, these metrics are predictive performance-oriented. In this paper, we contribute to this line of fairness work in education by investigating the possibility and value of a fairness metric that accounts for the behaviors of the classifiers.

## 3. AN ALGORITHMIC FAIRNESS ANALYSIS APPROACH

### 3.1 Definition of the MADD metric

We introduce a novel metric, the Model Absolute Density Distance (MADD), which is based on measuring algorithmic fairness via models' behavior differences between two groups, instead of via models' predictive performance. It is worth noting that this focus on the models' behaviors enables us to not only quantify algorithmic fairness, but also gain a deeper understanding of how the models discriminate via graphical representations of the MADD (see next subsection 3.2).

We present the MADD under the scope of this study where we consider binary sensitive features and binary classifiers that output probability estimates (or confidence scores) associated to their predictions.

Assume a model $\mathcal{M}$, trained on a dataset $\{X, S, Y\}_{i=1}^{n}$ where $S$ are the binary sensitive features, $X$ all the other features characterizing the students, $Y \in \{0, 1\}$ the binary target variable, and $n$ the number of samples. $\{X, S\}_{i=1}^{n}$ represents all the features of the prediction task. More precisely, $S = (s_i^a)_{i=1}^n$ where $a$ is the index of the considered sensitive feature and $s_i^a \in \{0, 1\}$. Indeed, if a student $(x_i, s_i^a)$ belongs to any group named $G_0$ of the sensitive feature $a$, then $s_i^a = 0$, and idem $s_i^a = 1$ if $(x_i, s_i^a)$ belongs to the other group named $G_1$ of the same sensitive feature $a$. Note that a sample $(x_i, s_i^a)$ describes a unique student in a group, with the groups $G_0$ and $G_1$ being mutually exclusive (i.e. a student can only belongs to one of these two groups). Also, none of these groups is considered as a baseline or privileged group here.

$\mathcal{M}$ aims at minimizing some loss function $\mathcal{L}(Y, \hat{Y})$ with its predictions $\hat{Y}$ to estimate or predict $Y$. $\mathcal{M}$ should assign to each $\hat{Y}_i$ a predicted probability (or a confidence score) that a given sample $(x_i, s_i^a)$ will be predicted as $\hat{Y}_i = 1$. This probability or score is noted $\hat{p}(x_i, s_i^a) = P(Y = 1|X_i = x_i, S = s_i^a)$. We introduce a parameter $e$ that is the probability sampling step of $\hat{p}$ values between 0 and 1. In other words, $\hat{p}$ values are rounded to the nearest $e$ (e.g. $\hat{p}(x_i, s_i^a) = 0.09$ if $e = 0.01$ and the same $\hat{p}(x_i, s_i^a) = 0.1$ if $e = 0.1$ for instance). $\mathcal{M}$ predicts $\hat{Y}_i = 1$ if and only if $\hat{p}(x_i, s_i^a) \geq t$ where $t$ is a probability threshold, and $\hat{Y}_i = 0$ otherwise.

We define two unidimensional vectors $D_{G_0}^a$ and $D_{G_1}^a$ as what we call in short the *density vectors* of the respective groups $G_0$ and $G_1$ of the sensitive feature $a$. They actually contain all the density values associated to each $\hat{p}(x_i, s_i^a)$ value (rounded to the nearest $e$) of group $G_0$ or group $G_1$. In particular, $D_{G_0}^a = (d_{G_0,k}^a)_{k=0}^m$ where each $d_{G_0,k}^a$ is the density of $\hat{p}(x_i, s_i^a) = k \times e$ value, that is to say the frequency that the model $\mathcal{M}$ gives $\hat{p}(x_i, s_i^a) = k \times e$ divided by the sum of frequency of all $\hat{p}$ values. $m$ is equal to the total number of distinct $\hat{p}$ values and is related to $e$ by the following: $m = 1/e + 1$. The advantage of the introduction of $e$ could be seen here: having discretized the $\hat{p}$ values enables us to have the two density vectors $D_{G_0}^a$ and $D_{G_1}^a$ of the same length so that they are comparable on the probability space and independent from the model $\mathcal{M}$'s behaviors.

We now define the MADD as follows:

$$\text{MADD}(D_{G_0}^a, D_{G_1}^a) = \sum_{k=0}^{m} |d_{G_0,k}^a - d_{G_1,k}^a| \qquad (1)$$

The MADD satisfies the necessary properties of a metric: reflexivity, non-negativity, commutativity, and triangle inequality [9] (see the proofs in Appendix A). Moreover:

$$\forall a, \quad 0 \leq \text{MADD}(D_{G_0}^a, D_{G_1}^a) \leq 2 \qquad (2)$$

The closer the MADD is to 0, the fairer the outcome of the model is regarding the two groups. Indeed, if the model produces the same probability outcomes for both groups, then $D_{G_0}^a = D_{G_1}^a$ and $\text{MADD}(D_{G_0}^a, D_{G_0}^a) = 0$. Conversely, in the most unfair case, where the model produces totally distinct probability outcomes for both groups, the MADD is equal to 2. An example of such a situation is when $\exists k_{G_0}, d_{G_0,k_{G_0}}^a = 1$ and $\forall k \in [0, m], k \neq k_{G_0}, d_{G_0,k}^a = 0$, and $\exists k_{G_1} \neq k_{G_0}, d_{G_1,k_{G_1}}^a = 1$ and $\forall k \in [0, m], k \neq k_{G_1}, d_{G_1,k}^a = 0$. In that case, Equation 1 becomes:

$$\text{MADD}(D_{G_0}^a, D_{G_1}^a) = |d_{G_0,k_{G_0}}^a| + |d_{G_1,k_{G_1}}^a| = (1 + 1) = 2 \qquad (3)$$

### 3.2 Visualization-based analysis of models' discriminatory behaviors

We introduce a visualization-based analysis of the models' discriminatory behaviors that complements our fairness analysis approach. This analysis is based on graphical interpretations of the MADD. Let us plot in Figures 1a and 1b the density histograms associated with each density vector $D_{G_0}^a$ and $D_{G_1}^a$. These histograms represent the distributions of the $\hat{p}$ values for the group $G_0$ and the group $G_1$ of a sensitive feature $a$. The number and consequently the width of the intervals depend on the probability sampling step $e$.

However, these histograms are not easily interpretable because of the numerous variations of the discrete values. We solve this issue by applying a smoothing by kernel density estimation (KDE) with Gaussian kernels, as shown in Figure 1c. The smoothing parameter, also called bandwidth parameter, is determined by the Scott's rule, an automatic bandwidth selection method[2]. This smoothing transforms the discrete probability distribution (whose density values cannot exceed 1 in the y-axis as they are related to discrete random variables) into a continuous approximation of the associated probability density function (PDF), which can in turn take values greater than 1.

Therefore, a visual approximation of the MADD corresponds to the red area in Figure 1d. Indeed, as the MADD uses the absolute density distances point-by-point between the two density vectors, the metric can be visually approximated by the area in-between the two curves, considering that the graph shows continuous density instead of the true discrete values used in the MADD calculation. Conversely, the green area, which is the intersection of the smoothed representations of the two density vectors, illustrates the area where the model $\mathcal{M}$ produces the same predicted probabilities for both groups up to a certain approximated density. We call this area the *fair zone*.

---

[2] See documentation of `scipy.stats.gaussian_kde`.

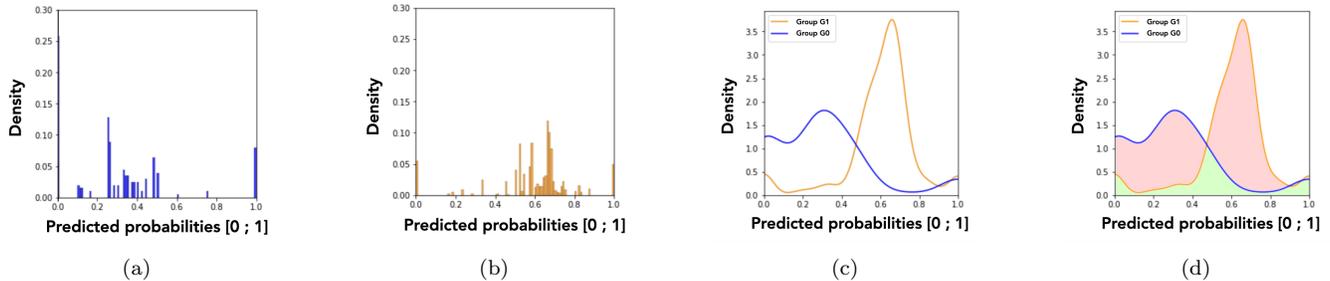

Figure 1: Visual representation of the MADD. Histograms of predicted probabilities for group $G_0$ (a) and group $G_1$ (b). Smoothing of these histograms (c). Approximation of the MADD in the red zone (d) vs. the *fair zone* in green.

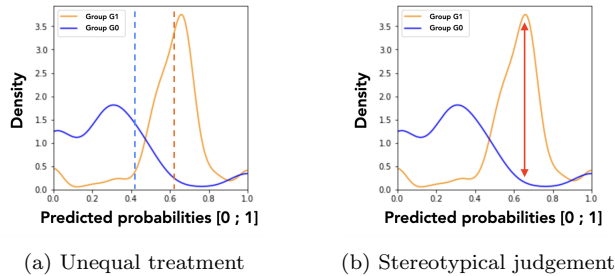

(a) Unequal treatment    (b) Stereotypical judgement

Figure 2: Two models' discriminatory behaviors. The dotted lines are the respective means of the two density vectors.

Thanks to this graphical representation approximating the MADD, we are able to distinguish two model's discriminatory behaviors: unequal treatment (Figure 2a), and stereotypical judgement (Figure 2b). Unequal treatment behavior can be summarized as follows: "how much the model favor or penalize individuals based on them belonging to each group?" As displayed in Figure 2a, we can identify which group get lower or higher predicted probabilities on average, allowing us to understand which group the model tends to favor (the highest mean, here the group $G_1$) or to penalize (the lowest mean, here the group $G_0$). It is worth noticing that the means are not perfectly aligned with the peaks of the distributions because they are calculated from the density vectors, without the smoothing. The second discriminatory behavior, stereotypical judgement, can be summarized as follows: "how much the model makes repetitive and invariant "judgement" about the individuals based on them belonging to a group?" For instance in Figure 2b, the model clearly tends to give to many persons in the group $G_1$ the same predicted probabilities. These analyses cannot be performed with existing predictive performance-oriented fairness metrics, as the model could have the same accuracy for both groups regardless of its underlying effective predictions, either in terms of distributions or in terms of density differences.

## 4. EXPERIMENTAL SETTING

We apply our approach on the common task of predicting student success to a course, and we present in this section (1) the data, (2) the models, and (3) the setting parameters we used in our experiments. This case study is designed to further investigate our proposed approach, and to show how one can use it.

### 4.1 Data

#### 4.1.1 Dataset presentation

We used real-world anonymized data from the Open University Learning Analytics Dataset (OULAD) [22]. The Open University is a distance learning university from the United Kingdom, offering higher education courses which can be taken as standalone courses or as part of a university program with no previous qualifications required. The dataset contains both student demographic data and interaction data with the university's virtual learning environment (VLE). The students were enrolled in at least one of the three courses in Social Sciences or one of the four Science, Technology, Engineering and Mathematics (STEM) courses between 2013 and 2014. The dataset contains 32,593 samples including 28,785 unique students.

The choice of this dataset was motivated by several reasons. First, the OULAD is one of the most comprehensive and benchmark datasets in the learning analytics domain to assess the performance of students in a VLE [1]. In addition, it is an open dataset that answers the call to the community for the development of new approaches on open datasets [15]. Then, it also answers another call from [15] for replication in multiple contexts such as several courses with diverse populations, as provided in the OULAD. Moreover, as it is commonly the case with distance learning universities, the students have a large variety of profiles [8] (including on average more women than men and a wide age range [2]), and these information are available in the dataset, making it particularly relevant for studying the impact of demographic features in terms of fairness. Finally, the data was collected in compliance with The Open University requirements regarding ethics and privacy, including consent and anonymization.

#### 4.1.2 Data preprocessing

We used the features presented in Table 1. The `sum_click` feature was the only one that was not immediately available in the original dataset and was computed from inner joints and aggregation on the original data. Also, we removed samples where the value of the `poverty` feature was missing (4% of the data samples) and when the students withdrew from the courses (24% of the data samples). This left us with 19,964 samples of distinct students, whose values were scaled between 0 and 1 for every feature via normalization. We indeed did not apply standardization to keep the original data distributions and analyze the models' behaviors accordingly. The target variable (course outcome) was coded as "Pass" or "Fail" (1 or 0 respectively). Plus, students who got a "Distinction" outcome were also coded as 1 ("Pass"), as we target binary classification for this case study.

Table 1: Features used from the OULAD dataset [22].

| Name | Feature type | Description |
| --- | --- | --- |
| `gender` | binary | the students' gender |
| `age` | ordinal | the interval of the students' age |
| `disability` | binary | indicates whether the students have declared a disability |
| `highest_education` | ordinal | the highest student education level on entry to the course |
| `poverty`[3] | ordinal | specifies the Index of Multiple Deprivation [22] band of the place where the students lived during the course |
| `num_of_prev_attempts` | numerical | the number of times the students have attempted the course |
| `studied_credits` | numerical | the total number of credits for the course the students are currently studying |
| `sum_click` | numerical | the total number of times the students interacted with the material of the course |

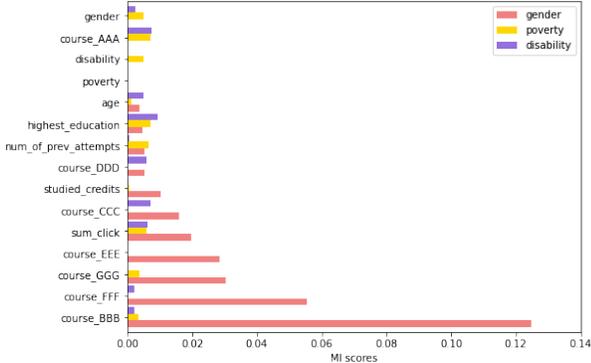

Figure 3: Mutual information (MI) scores.

In our study, we considered three sensitive features: `gender`, `poverty`, and `disability`. Although other sensitive features could have been relevant, our main focus here is in investigating our proposed method itself. Therefore, one may choose different sensitive features according to their purpose (for instance including age as well, as in [34]), and one would be able to conduct the same fairness analysis process. Due to our method dealing with binary features as sensitive features, we transformed `poverty` into a binary feature by setting a 50% threshold of deprivation index [22], coding as 0 those below the 50% threshold (i.e. less deprived) and as 1 those above (i.e. more deprived).

We did not apply any data balancing techniques nor unfairness mitigation preprocessing, still to keep the original data distributions. However, our approach does not prevent the use of such preprocessing.

### 4.1.3 Data analyses and course selection

We explored the correlations and imbalances of the sensitive features across the different courses in the dataset to identify those which were relevant for analysing algorithmic fairness. We thus computed the mutual information (MI) between all the features and the three sensitive ones, whose respective results are distinguished by a different color as shown in Figure 3. Mutual information is particularly relevant for non-linear relationships between features. Figure 3 shows that the course "BBB" followed by the course "FFF" are the most correlated with the `gender` feature, with the overall highest MI scores. Therefore, the Social Sciences course coded as "BBB" and the STEM course coded as "FFF", two different student populations, were good candidates for examining the impact of gender bias on the predictive models fairness. In addition, both courses presented very high imbalances in terms of `disability` (respectively 91.2-8.8% and 91.7-8.3% for 0-1 groups in courses "BBB" and "FFF") and `gender` (respectively 88.4-11.6% and 17.8-82.2% for 0-1 groups in courses "BBB" and "FFF"), and still some imbalance for `poverty` (respectively 42.3-57.7% and 46.9-53.1% for 0-1 groups in courses "BBB" and "FFF"). Based on these preliminary unfairness expectations derived from the skews in the data, it is interesting to analyze whether and how the models will suffer from these biases in both courses. These two courses are thus excellent testbeds for testing our approach.

## 4.2 Classification Models

To show that our fairness analysis approach can handle several types of classification models, we chose models either based on regression, distance, trees, or probabilities. More precisely, we chose a logistic regression classifier (LR), a k-nearest neighbors classifier (KN), a decision tree classifier (DT), and a naive bayes classifier (NB).

We chose these particular models for the following reasons. Firstly, they are widely used in education, and specially with the OULAD [21, 1]. Models based on vectors (e.g. support vector machines), also commonly used, were not selected as they do not outcome probability estimates (or confidence scores) on which to run our fairness analysis. Secondly, while our approach can be generalized to other models with probability estimates (or confidence scores) such as random forest or neural networks, we favored white boxes and explainability over finding the best modeling with fine-tuning. Thirdly, predicting students' success with the data in the OULAD is a rather low abstraction task due to the small amount of features and variance in the data, for which using complex predictive models would not lead to better performance and could even overfit the data. Finally, the selected models are easy to implement for most use cases, which makes them universally good candidates for predictive modeling in general.

---
[3]Named as `imd_band` in the original data.

To fit the models, we split the data into a train and a test set using a 70-30% split ratio in a stratified way, meaning that we kept the same proportion of students who passed and failed in both the train and the test sets. The resulting accuracies of the models were above the baseline (70%) and up to 93%, except for the NB (62%) which instead presented interesting behaviors with the MADD analyses and was worth keeping it (see Section 5). It has to be noted that, contrary to most ML studies, achieving the best predictive performance was not our focus here, since the purpose of our experiments is rather to analyze the fairness of diverse models with the MADD metric. Then, we used the models' outcomes on the test set to compute the MADD metric and generate the visualizations.

## 4.3 Fairness Parameters

For our study, we set $e$ to 0.01 (i.e. $m = 101$), and $t$, the probability classification threshold, to 0.5. For $e$, 0.01 corresponds to a variation of the probability of success or failure of 1%, which we deem a sufficient level of probability sampling precision, considering on the one hand that probability variations below 1% are not significant enough in the problem, and on the other hand that higher values of $e$ (up to 0.1) did not alter the MADD results. Regarding $t$, the success prediction is generally defined by having an average score above 50% and thus we chose $t$ with respect to the problem rather than optimizing it for model performance. The odds of positive or negative predictions are thus balanced and the threshold is the same for each individual.

## 5. RESULTS

In the following, we show in subsection 5.1 how the MADD and its visualization-based analysis can help unveil unexpected results based on (1) the respective importance of each sensitive feature in algorithmic unfairness, (2) the models intrinsic unfairness, and (3) the nature of the unfairness associated with the predictions made by the model. Then, we show in subsection 5.2 how our results differ from and complement what can be provided by ABROCA, a state-of-the-art predictive performance-oriented fairness metric. Both subsections 5.1 and 5.2 are concluded by a summary of the obtained results.

## 5.1 Fairness Analysis with MADD

In the parts 5.1.1 and 5.1.2, we examine via Tables 2 and 3 the MADD results reported for the two courses. We highlight in bold the best MADD per column, and with an asterisk (*) the best MADD per row. In this way, the MADD of the fairest model for each sensitive feature is in bold, whereas the MADD of the fairest sensitive feature for each classifier is marked with a *. As examples, in Table 2 the DT is the fairest model regarding the `poverty` feature (bold), and in Table 3 the `disability` feature is the fairest for the KN (*). For the part 5.1.3, we base our visual analyses and identification of discriminatory behaviors explained in subsection 3.2 on Figures 4 and 5.

### 5.1.1 Sensitive features analysis

*Course "BBB" (Social Sciences).* Table 2 reveals that three models out of four (LR, KN, and DT) are the fairest for the `disability` sensitive feature. Therefore, two interesting observations can be made. First, it is contrary to what we would expect since `disability` was the most imbalanced (91.2-8.8% for 0-1 groups) sensitive feature in the training data (see Section 4.1.3). Second, the `gender` feature was particularly expected to be highly sensitive due to its high correlation with the target in this course and its imbalance, but it actually has the best MADD on average (1.02).

*Course "FFF" (STEM).* Similarly to the above results for the course "BBB", we can notice that the data skews are not necessarily reflected in the MADDs. In the training data, the `disability` sensitive feature was highly imbalanced, and the `poverty` feature was quite balanced. Nonetheless, for half of the models (see Table 3), both `disability` and `poverty` are the two sensitive features with regard to which the models are the fairest. On the other hand, in line with the gender skew and correlation shown in Section 4.1.3, `gender` has the worst MADD results in average, more than `disability`, although the difference is not substantial.

### 5.1.2 Model fairness analysis

*Course "BBB" (Social Sciences).* Now focusing on the fairness of the models, DT appears in Table 2 to be the fairest, with an average MADD of 0.73 across all the sensitive features. DT is indeed the fairest for `disability` and `poverty` and the second best for `gender`. On the contrary, LR is the least fair, with the highest results for each senstive feature and an average of 1.71, with a maximum value of 2.

*Course "FFF" (STEM).* NB and DT obtain the best MADD averages across all three sensitive features (0.64 and 0.65 respectively). Therefore, there is no clear winner for this course as they behave differently according to different sensitive features: NB has better results for `gender` and `poverty` but a higher MADD for `disability`, whereas DT is more balanced across the three sensitive features. However, we remind that NB performed below the accuracy baseline and thus DT would overall be a better candidate.

Table 2: MADD results for the course "BBB".

|      | Model | Sensitive features |         |            | Average |
|------|-------|--------|---------|------------|---------|
|      |       | gender | poverty | disability |         |
| MADD | LR    | 1.72   | 1.85    | 1.57*      | 1.71    |
|      | KN    | 1.13   | 1.12    | 0.93*      | 1.06    |
|      | DT    | 0.69   | **0.85**| **0.65***  | 0.73    |
|      | NB    | **0.52*** | 0.9  | 1.37       | 0.93    |
| Average |    | 1.02   | 1.18    | 1.13       |         |

Table 3: MADD results for the course "FFF".

|      | Model | Sensitive features |         |            | Average |
|------|-------|--------|---------|------------|---------|
|      |       | gender | poverty | disability |         |
| MADD | LR    | 1.18   | 1.06*   | 1.12       | 1.12    |
|      | KN    | 1.06   | 0.93    | 0.78*      | 0.92    |
|      | DT    | 0.76   | 0.65    | **0.55***  | 0.65    |
|      | NB    | **0.56** | **0.47*** | 0.90   | 0.64    |
| Average |    | 0.89   | 0.78    | 0.84       |         |

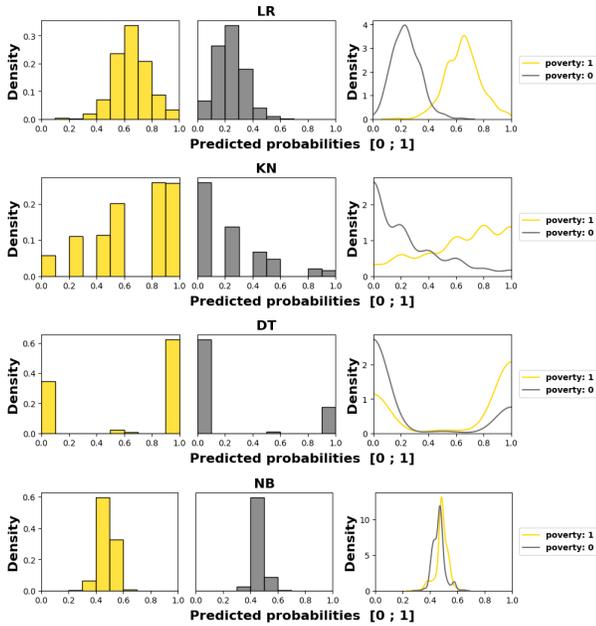

Figure 4: Models' behaviors in course "BBB". Note that for these graphs $e$ was set to 0.1 for better visualization of the bars, but $e$ was actually equal to 0.01 for computation, as said in subsection 4.3.

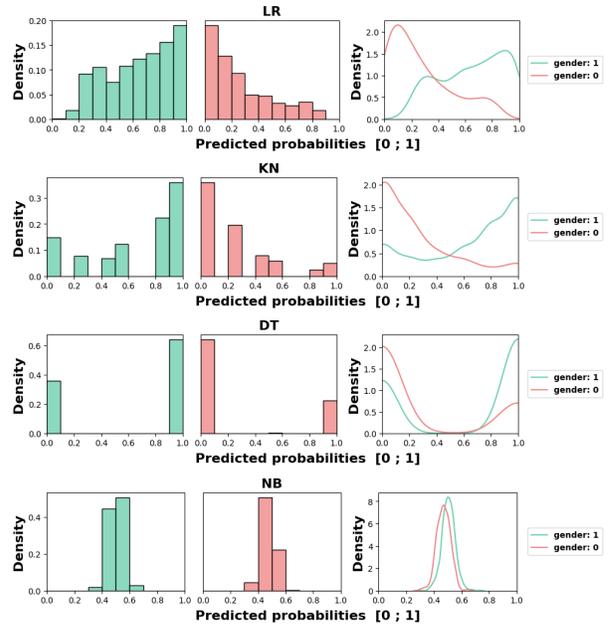

Figure 5: Models' behaviors in course "FFF". Note that for these graphs $e$ was set to 0.1 for better visualization of the bars, but $e$ was actually equal to 0.01 for computation, as said in subsection 4.3.

#### 5.1.3 Visualization-based analysis

*Course "BBB" (Social Sciences).* We examine in Figure 4 the models' behaviors regarding the most sensitive feature, namely `poverty` in this course, as it has the worst MADD on average (1.18). We see in the subfigures, through an offset of the distribution mean to the left for the group 0, that three models out of four (LR, KN, and DT) have learned unequal treatment against those in better financial conditions (group 0). Among them, KN and DT present the highest stereotypical results reduced to only few probability values, which illustrates well their inner workings. Conversely, NB produces the least discriminating results with the closest means for the two groups. Its behavior is all the more interesting since it shows that having poor predictive performance is not necessarily interfering with behaving fairly regarding two groups of the most sensitive feature. It is precisely because it does not discriminate against any features, whether they were sensitive or not, that it has poor accuracy.

*Course "FFF" (STEM).* Likewise, we examine in Figure 5 the models' behaviors for the most sensitive feature in this course, `gender`, with the worst MADD average of 0.89. All models but NB exhibit unequal treatment against group 0, here the women. Similarly to the previous results, we can again note the highly stereotypical behaviors of KN and DT, and the relative fairness of the NB model.

#### 5.1.4 Implications for the MADD

Following the double reading of the tables, feature-wise or model-wise, as well as our visual analyses, we can make two important observations regarding the insights provided by the MADD.

Firstly, there is no direct relationships between biases in the data (imbalanced representations, high correlations) and the discriminatory behaviors learned by the models. We even observe opposite conclusions (specially for the course "BBB" in part 5.1.1).

Secondly, trained on the same data, the models exhibit very different discriminatory behaviors (see parts 5.1.2 and 5.1.3), both regarding different sensitive features, and different severity and nature of their algorithmic unfairness. This was also shown by our visual analysis, which allowed finer-grained interpretations of the discriminatory behaviors.

### 5.2 Comparison with ABROCA

We now aim to compare the MADD with the ABROCA predictive performance-oriented fairness metric [14]. The ABROCA results (computed with the source code from [12]) are displayed in Tables 4 and 5, and an illustrated example for the course "BBB" is given through the Figures 6 and 7.

#### 5.2.1 Sensitive features analysis

*Course "BBB" (Social Sciences).* Let us first focus on the `poverty` feature which has the worst MADD average (1.18 from Table 2). In particular, in part 5.1.1, `poverty` was the feature with which LR obtained the worst MADD (1.85 from Table 2), which was also the worst MADD overall. Indeed, in Figure 6 it can be seen that LR has the smallest intersection area compared to the other models. However, in Figure 7 and Table 4 we see that LR has one of the best

ABROCA (0.03) with minimal area between the curves of the respective groups. We found similar opposite results between MADD and ABROCA for `gender`. Thus, `poverty` and `gender` could be seen as unfair sensitive features for a model on the one hand (MADD) and as fair ones on the other hand (ABROCA). Moreover, DT too has one of the best ABROCA (0.03), while it provided the best MADD value (0.85 from Table 2) regarding this feature. Therefore, two models with the same ABROCA lead to opposite discriminatory behaviors according to the MADD. In the end, ABROCA and MADD do not highlight the same fairness results, and can even lead them to show opposite results.

*Course "FFF" (STEM).* Now examining Table 5 for the course "FFF", ABROCA does not capture substantial differences at the sensitive feature level (column-wise), with an ABROCA average of 0.4 for all three features. Thus, the MADD results can capture additional differences among the models' behaviors that are not reflected in the ABROCA results. In addition, we again found that `disability`, the most imbalanced sensitive feature, is actually the feature for which DT is the fairest, regardless of whether we consider the MADD in part 5.1.1 (Table 3) or the ABROCA (Table 5). Therefore, ABROCA does not reflect the imbalance bias in the data either, in contradiction with the findings from [14].

### 5.2.2 Model fairness analysis

*Course "BBB" (Social Sciences).* In Table 4, LR and NB appear to be the fairest models across all the sensitive features (best common ABROCA average of 0.3). However, with the MADD, NB indeed exhibited overall quite balanced low values, but LR was always the least fair on average (Tables 2 and 3). Thus, at the model level this time, the trends in the MADD results are only partially reflected in the ABROCA results.

*Course "FFF" (STEM).* Table 5 shows that ABROCA results do not exhibit substantial variability to distinguish differences in fairness between the models (row-wise this time) in our experiment. In addition to similar ABROCA averages, all models have very close ABROCA results specially regarding the `gender` and `poverty` features. Therefore, the MADD allows to find complementary discriminatory results as compared to using only ABROCA.

### 5.2.3 Summary of the comparison

Two main takeaways could be reported from our comparison between the ABROCA and MADD metrics.

Firstly, fair predictive performance (i.e. similar numbers of errors across groups, here captured by low ABROCA values) does not guarantee fair models' behaviors (i.e. low severity of discrimination across groups, here captured by low MADD values). This demonstrates what we advocated in the introduction (Section 1) regarding investigating the models' behaviors to gain a comprehensive understanding of the models fairness. In particular, two models with the same ABROCA could suffer from substantial, and even op-posite algorithmic discriminatory behaviors, which can be uncovered by the MADD (see parts 5.2.1 and 5.2.2). Using the MADD together with a predictive performance-oriented metric such as ABROCA can thus allow more informed selection of fair models in education, and here in our experiments, they provide strong evidence that DT is the fairest model on both courses.

Table 4: ABROCA results for course "BBB".

| Model | Sensitive features | | | Average |
|---|---|---|---|---|
| | gender | poverty | disability | |
| ABROCA LR | 0.02 | 0.03 | 0.03 | 0.03 |
| ABROCA KN | 0.08 | 0.06 | 0.06 | 0.07 |
| ABROCA DT | 0.06 | 0.03 | 0.05 | 0.05 |
| ABROCA NB | 0.04 | 0.02 | 0.04 | 0.03 |
| Average | 0.05 | 0.04 | 0.05 | |

Table 5: ABROCA results for course "FFF".

| Model | Sensitive features | | | Average |
|---|---|---|---|---|
| | gender | poverty | disability | |
| ABROCA LR | 0.04 | 0.03 | 0.03 | 0.03 |
| ABROCA KN | 0.04 | 0.05 | 0.04 | 0.04 |
| ABROCA DT | 0.05 | 0.04 | 0.01 | 0.03 |
| ABROCA NB | 0.03 | 0.03 | 0.07 | 0.04 |
| Average | 0.04 | 0.04 | 0.04 | |

Secondly, in line with our previous findings that biases in the data may not be related with models' discriminatory behaviors (see part 5.1.4), we also observed that the biases in the data are independent from predictive performance biases too. For instance, the highest imbalanced sensitive feature could actually lead to both the best ABROCA and the best MADD. Although this observation is aligned with the findings in [11, 17], it is worth noting that it goes against what the authors of ABROCA had observed [14] (see part 5.2.1).

## 6. DISCUSSION

In this section, we discuss (1) the overall implications of the results of our fairness study, (2) the limitations and the strengths of the proposed approach, (3) some potential experimental improvements, and (4) some guidelines to use our fairness analysis approach with the MADD.

### 6.1 Fairness results

Our results lead to three main conclusions, as follows. Firstly, we found no direct relationships between data bias and predictive performance bias nor discriminatory behaviors bias. It confirms previous findings that unfair biases are not only captured in the data, but are inherent to the model too [27, 28]. It further suggests that exclusively mitigating unfairness in the data might not be sufficient, and that mitigating unfairness at the model level is key too. Secondly, even trained on the same data, each model exhibits its own discriminatory behavior (likely linked to its inner working) and according to different sensitive features. It raises interesting questions on how different models could be combined in order to balance discriminatory behaviors with regards to multiple sensitive features at the same time. Thirdly, fair predictive performance does not guarantee by itself fair models' behaviors and thus fair outcomes. Additional introspection of the model is therefore needed, and our approach

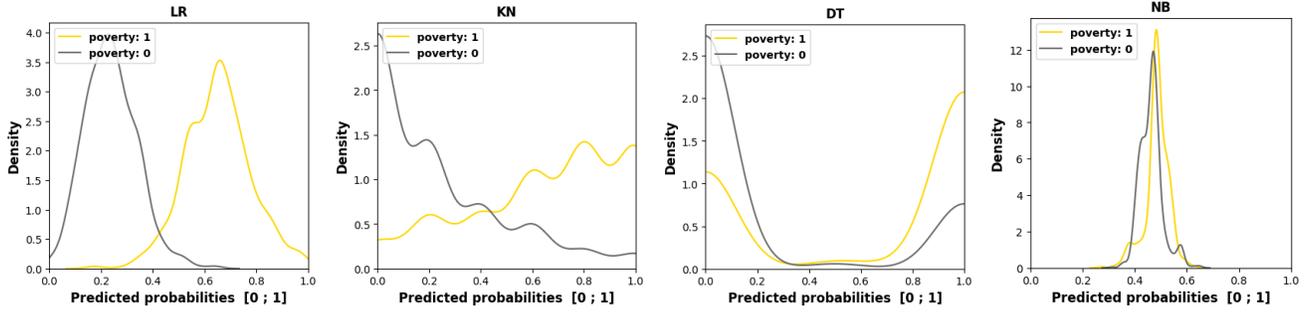

Figure 6: MADD visualizations for the `poverty` sensitive feature across all the models for course "BBB".

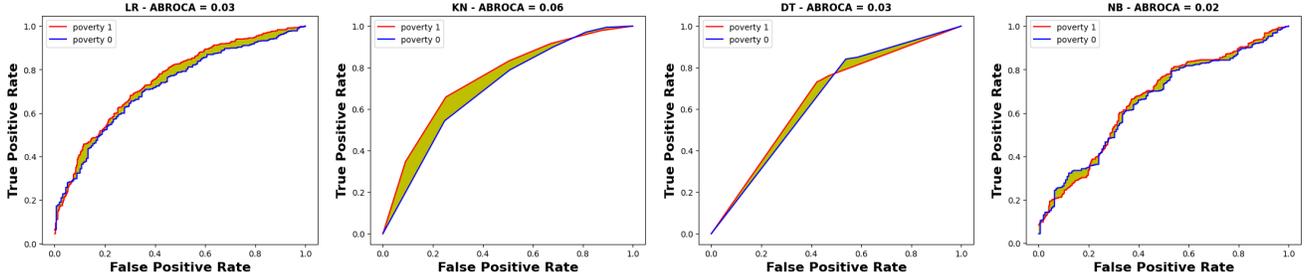

Figure 7: ABROCA slide plots for the `poverty` sensitive feature across all the models for course "BBB".

appears as a possible solution.

## 6.2 Limitations and strengths

Although our approach was initially designed for analyzing algorithmic fairness at an individual sensitive feature level, our prospective work includes a generalization of the MADD metric to capture the influence of multiple sensitive features simultaneously. Moreover, the current MADD is particularly suitable for binary sensitive features and binary classifiers, and future work should also focus on extending it to multi-class features and classifiers. As an example, an extension for categorical sensitive features would enable us to have a finer-grained analysis of discrimination across more relevant subgroups. Despite these current limitations, the strengths of our approach stand in (1) its ability to be used with any tabular data, the most prevalent data representation [24] from any domain, and without needing any unfairness mitigation preprocessing; (2) being able to have a richer understanding of models' discriminatory behaviors and their quantification with an easy-to-implement fairness metric that is independent from predictive performance; and (3) since the MADD is bounded, being comparable between different datasets to measure the discriminatory influence of a particular sensitive feature in different contexts and populations.

## 6.3 Experimental improvements

In our experiments we purposely focused on the MADD results to highlight its contribution and interest to fairness analysis, however for real-case applications one should obviously pay attention to both predictive performance and fairness performance in order to thoroughly select satisfying models. As an example, the NB model used in our experiments could be seen as fair regarding its MADD results, but it had in fact poor accuracy particularly because it was unable to predict well the success or the failure of students regarding any features, which makes this model not usable for real-case purposes but nonetheless interesting for our exploratory analysis. We thus recommend using the MADD on models that show satisfying predictive performance, to gain a finer-grained understanding on how they behave and regarding who and to refine models selection and their usage. Moreover, one should also consider testing variations of the probability sampling parameter $e$ in their application and context. Although the impact of its variation in the range from 0.01 to 0.1 was low in our experiments, it might not be always the case. Determining the optimal value for this parameter is also a key part of our prospective work. Finally, we have demonstrated the validity and value of our approach on two courses of the OULAD dataset. Nonetheless, in a broader context of investigating model unfairness, this work should be replicated with other educational datasets providing more students data and more diverse sensitive features [3].

## 6.4 Guidelines

In order to facilitate replication studies and the use of our approach (in addition to the availability of the data and our source code), we provide in the following a 7-step guide to help readers compute the MADD and plot the models' behaviors as in Figures 4 and 5.

1. Choose binary classification models that can output probability estimates or confidence scores.

2. Transform, when needed, every sensitive feature into binary one.

3. Train the models, and in the testing phase separate

their predicted probabilities or confidence scores according to the groups of each sensitive feature.

4. Compute the MADD for each sensitive feature, and compare the results between features and models.

5. Plot histograms of the predicted probability distributions of each group of the sensitive features, and their smoothed estimations (e.g. with KDE).

6. Visually identify discriminatory behaviors among unequal treatment (i.e. distance between the two distribution means) and stereotypical judgement (i.e. differences of local amplitudes).

7. Depending on the fairness analysis goals:
    - Identify which models are the fairest overall or according to which sensitive features, using a row-wise reading of the results table.
    - Identify which features are the most sensitive overall or according to which models, using a column-wise reading of the results table.
    - Using the plots, identify which groups (i.e. which distributions) are the most discriminated against by the models (relatively to each sensitive feature).

## 7. CONCLUSION

In this paper, we developed an algorithmic fairness analysis approach based on a novel metric, the *Model Absolute Density Distance* (MADD). It measures models' discriminatory behaviors between groups, independently from their predictive performance. Our results on the OULAD dataset and comparison with ABROCA show that (1) fair predictive performance does not guarantee fair models' behaviors and thus fair outcomes, (2) there is no direct relationships between data bias and predictive performance bias nor discriminatory behaviors' bias, and (3) trained on the same data, models exhibit different discriminatory behaviors and according to different sensitive features.

This approach, for which we provide a set of guidelines in subsection 6.4 and our source code and data in open access at https://github.com/melinaverger/MADD, can be used to help identify fair models, exhibit sensitive features, and determine students who were the most discriminated against and how (unequal treatment or stereotypical judgement) in an education context. Being bounded, an advantage of this metric is that it can be used across different contexts and data to discover the features that more generally cause algorithmic discrimination.

Future work will involve the generalization of the MADD metric to multiple sensitive features, its extension to multi-class sensitive features and classifiers, determining the optimal probability sampling parameter, and we will investigate how to use the MADD as an objective function to optimize models accordingly (in addition to predictive performance objectives).

## 8. ACKNOWLEDGMENTS
The authors want to thank Chunyang Fan (Sorbonne Université) for his valuable comments on this work.

# APPENDIX
## A. PROOFS FOR MADD AS A METRIC

We remind the definition of the MADD:

$$\text{MADD}(D^a_{G_0}, D^a_{G_1}) = \sum_{k=0}^{m} |d^a_{G_0,k} - d^a_{G_1,k}| \quad (1)$$

The MADD satisfies the necessary properties of a metric [9]:

$$\text{MADD}(D^a_{G_0}, D^a_{G_0}) = 0 \quad \text{reflexivity} \quad (4)$$

$$\text{MADD}(D^a_{G_0}, D^a_{G_1}) \geq 0 \quad \text{non-negativity} \quad (5)$$

$$\text{MADD}(D^a_{G_0}, D^a_{G_1}) = \text{MADD}(D^a_{G_1}, D^a_{G_0}) \quad \text{commutativity} \quad (6)$$

$$\text{MADD}(D^a_{G_0}, D^a_{G_2}) \leq \text{MADD}(D^a_{G_0}, D^a_{G_1}) + \text{MADD}(D^a_{G_1}, D^a_{G_2}) \quad \text{triangle inequality} \quad (7)$$

*Proof for reflexivity (Eq. 4)*

$$\text{MADD}(D^a_{G_0}, D^a_{G_0}) = \sum_{k=0}^{m} |d^a_{G_0,k} - d^a_{G_0,k}| = 0$$

*Proof for non-negativity (Eq. 5)*
Due to the positivity of each term in the sum thanks to the absolute value operator, the sum of these positive terms is always positive and $\text{MADD}(D^a_{G_0}, D^a_{G_1}) \geq 0$.

*Proof for commutativity (Eq. 6)*
Let $x$ and $y$ be real numbers. By commutativity of the absolute value operator, $|x - y| = |y - x|$. Thus, for any $k$, $|d^a_{G_0,k} - d^a_{G_1,k}| = |d^a_{G_1,k} - d^a_{G_0,k}|$ and then $\text{MADD}(D^a_{G_0}, D^a_{G_1}) = \text{MADD}(D^a_{G_1}, D^a_{G_0})$.

*Proof for triangle inequality (Eq. 7)*
Let $x$ and $y$ be real numbers. By triangle inequality of the absolute value operator, $|x+y| \leq |x| + |y|$. Let $x = d^a_{G_0,k} - d^a_{G_1,k}$ and $y = d^a_{G_1,k} - d^a_{G_2,k}$. Then, for any $k$:

$$|x+y| \leq |x| + |y|$$
$$\Leftrightarrow |(d^a_{G_0,k} - d^a_{G_1,k}) + (d^a_{G_1,k} - d^a_{G_2,k})| \leq |d^a_{G_0,k} - d^a_{G_1,k}| + |d^a_{G_1,k} - d^a_{G_2,k}|$$
$$\Leftrightarrow |d^a_{G_0,k} - d^a_{G_2,k}| \leq |d^a_{G_0,k} - d^a_{G_1,k}| + |d^a_{G_1,k} - d^a_{G_2,k}|$$

Then, by linearity of the sum :

$$\sum_{k=0}^{m} |d^a_{G_0,k} - d^a_{G_2,k}| \leq \sum_{k=0}^{m} |d^a_{G_0,k} - d^a_{G_1,k}| + \sum_{k=0}^{m} |d^a_{G_1,k} - d^a_{G_2,k}|$$
$$\Leftrightarrow \mathrm{MADD}(D^a_{G_0}, D^a_{G_2}) \leq \mathrm{MADD}(D^a_{G_0}, D^a_{G_1}) + \mathrm{MADD}(D^a_{G_1}, D^a_{G_2})$$